\documentclass{article}

\usepackage{arxiv}

\usepackage{times}

\usepackage{soul}
\usepackage{url}
\usepackage[hidelinks]{hyperref}
\usepackage[utf8]{inputenc}
\usepackage[small]{caption}
\usepackage{graphicx}
\usepackage{amsmath}
\usepackage{booktabs}
\usepackage{tabularx}
\usepackage{listings}
\usepackage{tcolorbox}
\usepackage[ruled,vlined]{algorithm2e}
\usepackage{nicefrac}       
\usepackage{microtype}      
\usepackage{amsthm}
\usepackage{amsfonts}
\usepackage{pythonhighlight}

\theoremstyle{definition}

\title{Imbalance-XGBoost: leveraging weighted and focal losses for binary label-imbalanced classification with XGBoost}

\author{
  Chen Wang \\
  Department of Computer Science\\
  Rutgers University - New Brunswick\\
  Piscataway, NJ 08854, USA \\
  \texttt{chen.wang.cs@rutgers.edu} \\
  \And
 Chengyuan Deng \\
  Department of Computer Science\\
  Rutgers University - New Brunswick\\
  Piscataway, NJ 08854, USA \\
  \texttt{charles.deng@rutgers.edu} \\
    \AND
  Suzhen Wang \\
  Department of Health Statistics\\
  Weifang Medical University\\
  Weifang, Shandong, 261053 China \\
  \texttt{wangsz@wfmc.edu.cn} \\
}

\begin{document}
\maketitle

\begin{abstract}
The paper presents \emph{Imbalance-XGBoost}, a Python package that combines the powerful XGBoost software with weighted and focal losses to tackle binary label-imbalanced classification tasks. Though a small-scale program in terms of size, the package is, to the best of our knowledge, the first of its kind which provides an integrated implementation for the two loss functions on XGBoost and brings a general-purpose extension to XGBoost for label-imbalanced scenarios. In this paper, the design and usage of the package are discussed and illustrated with examples. Furthermore, as the first- and second-order derivatives of the loss functions are essential for the implementations, the algebraic derivation is discussed and it can be deemed as a separate contribution. The performances of the methods implemented in the package are extensively evaluated on Parkinson's disease classification dataset, and multiple competitive performances are presented with the ROC and Precision-Recall (PR) curves. To further assert the superiority of the methods, the performances on four other benchmark datasets from the UCI machine learning repository are additionally reported. Given the scalable nature of XGBoost, the package has great potentials to be broadly applied to real-life binary classification tasks, which are usually of large-scale and label-imbalanced.
\end{abstract}


\section{Introduction}
\label{sec:introduction}
XGBoost is an advanced Gradient Tree Boosting-based software that can efficiently handle large-scale Machine Learning tasks (\cite{chen2016xgboost}). Merited by its performance superiority and affordable time and memory complexities, it has been widely applied to a variety of research fields since been proposed, ranging from cancer diagnosis (\cite{wang2017benchmark}) and medical record analysis (\cite{wang2018ppscore}) to credit risk assessment (\cite{chang2018riskassess}) and metagenomics (\cite{wassan2019metagenomics}). Also, because of its easy-to-use Python interface and explainable nature, it has become the de facto method-of-the-first-choice for a majority of data science problems. For instance, in the famous \textit{Kaggle} (\url{https://www.kaggle.com/competitions}) competitions, many winning teams built their models based on XGBoost and expressed positive views on the method and its variations (\cite{chen2016xgboost,omar2018kaggle,nielsen2016kaggle}). It could be tentatively predicted that in the near future, XGBoost and its variations will remain one of the most-applied methods in the data science community. \par
On the other hand, although XGBoost has achieved considerable success on both regression and classification problems, its performance often becomes subtle when a situation of label-imbalanced classification is brought up. The problem of classification under label-imbalanced situations exists commonly in practice, and most accuracy-driven 'vanilla' Machine Learning methods suffer performance decline if the ratio between labels becomes biased. For example, in cancer diagnosis, it is common to have 95\% of the patients without cancer and only 5\% with it. If the model simply predict everyone as `no cancer', then the accuracy is 95\%, which is remarkably high. However, missing to spot any cancer patient can lead to fatal consequences. \par
There have been mixed reports on the capabilities of XGBoost in handling label-imbalanced data. For example, \cite{zhao2018mRNA} demonstrated through their experiments that XGBoost can outperform other methods on skewed datasets, while the figures in \cite{ruisen2018bagging} suggested that vanilla XGBoost must be combined with other ensembling methods to achieve satisfactory results. Given the popularity of XGBoost and the fact that label-skewed data is, unfortunately, commonly encountered in practice, this performance decay will leave significant negative effects on related research and applications.\par
This paper introduces \emph{imbalance-XGBoost}, an XGBoost-based Python package addressing the label-imbalanced issue in the binary label regime by implementing weighted (cross-entropy) and focal losses on the boosting machine. Weighted cross-entropy loss is one of the simplest algorithm-level cost-sensitive methods (\cite{sun2009imbalancereview}) for learning imbalanced data. It follows the straightforward idea to increase the penalization of the misclassification of a certain class, and it has been widely applied to adjust vanilla machine learning algorithms to the label-imbalanced domain (\cite{huang2016learning}). In contrast, focal loss (\cite{lin2017focal}) is a relatively novel method originated from research in object detection. The idea of the method is to add a $(1-y_{j})^{\gamma}$ factor to the cross-entropy function (where $y_{j}$ is the prediction of $p(x=j)$), and this will reduce the importance of the well-classified data points. Comparing with weighted cross-entropy, focal loss enjoys a more robust parameter configuration as the method will work in our favor as long as $\gamma > 0$.  \par
To the best of our knowledge, there has not been significant publication discussing the implementation of the two losses on XGBoost previously. Existing studies on XGBoost under label-imbalanced scenarios usually adopt data-level approaches such as re-sampling (\cite{kabir2018resampling}) and/or cost-sensitive loss with non-trainable \textit{a priori} modifications (\cite{xia2017cost}). \cite{chen2017radar} mentioned weighted XGBoost in their work, but details regarding the implementation are not presented. A major challenge in applying the two loss functions to XGBoost is that to approximate the incremental learning objective, first- and second-order derivatives of the loss functions with respect to the predictions must be presented (One can refer to section \ref{sec:derivatives} for more details on this). And an algebraic contribution of this paper is the derivations and implementations of the derivatives that enable the two losses to be run with XGBoost. \par
The package is written in Python with hard dependency on XGBoost, Numpy (\cite{van2011numpy}), and Scikit-learn (\cite{pedregosa2011sklearn}). The losses are integrated into the XGBoost system by the customized loss framework of the software, provided the derived expressions of the derivatives. Since the major methods in the program are included in the dependency graph, the core part of the package is of small scale, with only a few hundred lines of codes. Nevertheless, the function derivatives and implementations and the significance in practical applications make the work non-trivial. The software has been released on Github\footnote[1]{\url{https://github.com/jhwjhw0123/Imbalance-XGBoost}} and PyPI\footnote[2]{\url{https://pypi.org/project/imbalance-xgboost/}}, and it has started to attract users with considerable downloads. \par
The rest of the paper goes as follows. Section \ref{sec:softwaredesign} introduces the package from the perspectives of toolkit designers and users; Section \ref{sec:derivatives} provides the theoretical foundation of the second-order approximation of gradient boosting trees used in XGBoost and the first- and second-order derivatives of the two losses; Some related studies are surveyed and discussed in section \ref{sec:literaturereview}, then the experimental performances of the package are empirically examined on five imbalanced binary classification datasets in section \ref{sec:experiment}; And finally, section \ref{sec:conclusion} gives a general conclusion of the paper.

\section{Design and Usage of Imbalance-XGBoost}
\label{sec:softwaredesign}
\subsection{Code Design}
\label{subsec:codedesign}
Though the XGBoost method has implementations in multiple languages, Python is picked as the language-of-choice for its wide recognition and application in data science. The codes follow the standard of \emph{PEP8}, and the project has been designed as open-source with codes on the Github page. We strive to keep the program consistent with 'standard' practices in Python-based data science. The input data is designated as \textit{Numpy array} (\cite{van2011numpy}), but by explicitly adding \texttt{np.array()} conversion, data types compatible with \textit{Numpy array} (e.g. \textit{Pandas Dataframe} (\cite{mckinney2011pandas})) will also work on the package. As a small project, the usage of it can be clearly presented with the Readme file, and there is no additional documentation required. \par
\begin{figure*}[!h]
\centering
\includegraphics[width=0.9\textwidth]{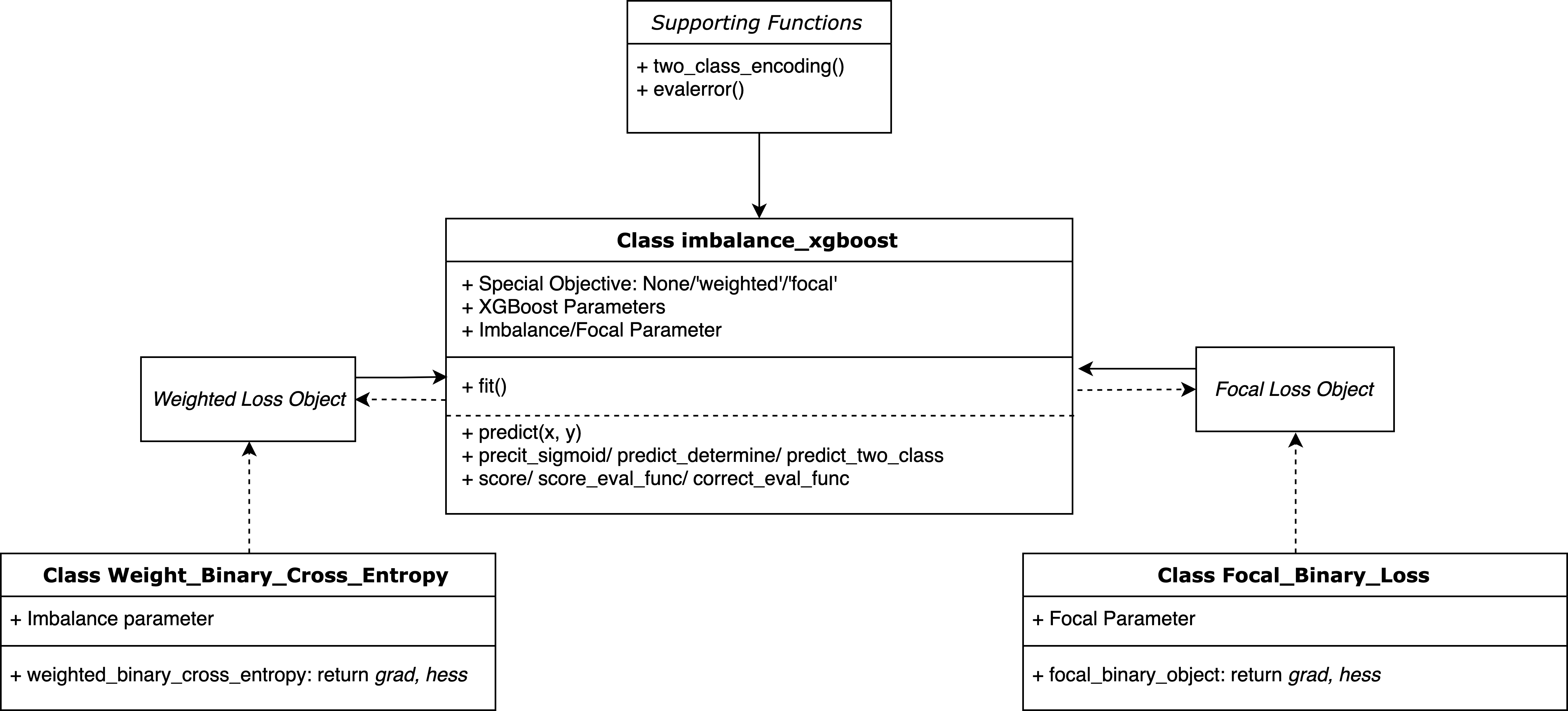}
\caption{The Overall Structure of the Program}
\label{fig:programstructure}
\end{figure*}

The overall program is consist of three classes: one main class \texttt{imbalance\_xgboost}, which contains the method the users will be applying, and two customized-loss classes, \texttt{Weight\_Binary\_Cross\_Entropy} and \texttt{Focal\_Binary\_Loss}, on which the imbalanced losses are based. The loss functions are designed as separate classes for the convenience of parameter tuning, and they are not supposed to be called by the users. When initializing an \texttt{imbalance\_xgboost} object, keyword \texttt{special\_objective} will be recorded by the \texttt{\_\_init\_\_()} method. Then, when executing the \texttt{fit()} function, the corresponding loss function object will be instantiated with the given parameter, and the built-in \texttt{xgb.train()} method of XGBoost will be able to fit the model based on the customized loss function. Figure \ref{fig:programstructure} illustrates the overall structure of the program. 

The package is designed to be an estimator class of the \textit{Scikit-learn} toolkit. This scheme enables the model and parameter selection methods in \textit{Scikit-learn}, such as \texttt{GridsearchCV()} and \texttt{RandomizedSearchCV()}, to be directly applied to find the best parameters for imbalanced XGBoosts. In practical data science tasks, this feature is crucial as the optimal models rely heavily on parameter tuning and selection. Also, estimator in \textit{Scikit-learn} can be combined with other estimators (transformers) by integrating them to a \texttt{Pipeline} object (\cite{pedregosa2011sklearn}). This allows the weighed- and focal-XGBoost to be easily combined with other pre-processing methods, such as resampling, to produce more robust results. Section \ref{subsec:usagesklearn} will provide more details for the package to tune parameters and perform cross-validation with \textit{Scikit-learn}. 
\par
\subsection{Model Optimization and Evaluation with Scikit-Learn}
\label{subsec:usagesklearn}
Similar to common classifiers in \textit{Scikit-learn}, the best classifier/parameter can be obtained through exhaustive or random search with the functions \texttt{GridsearchCV()}. It is noticeable that after fitting the model, it is possible to retrieve the 'plain' booster by accessing member \texttt{opt\_booster.booster}, and the object will be a XGBoost class (instead of Imbalance-XGBoost class). This makes it possible for the user to train the model on a machine where \textit{Imbalance-XGBoost} is available, save the model as 'plain' XGBoost, and run on another machine where only the original XGBoost package is installed. \par
Likewise, the cross-validation evaluation of \textit{Imbalance-XGBoost} is similar to the process on an 'ordinary' classifier, and one can simply follow the documentation of \textit{Scikit-Learn} to perform it. Notice that if one intends to perform a combination of parameter selection and cross-validation, it is recommended that a new booster with random initialization and selected parameters should be instantiated after the parameter selection, instead of directly using the fitted optimal booster from \texttt{GridsearchCV()}. The reason for doing so is that for the iterative training process of XGBoost, a booster trained from a randomized state is essential for a fair evaluation. 
\subsection{Built-in Evaluation Score}
\label{subsec:builtinscore}
There are three evaluation functions in the package. The overriding \texttt{score()} function serves the purpose to evaluate prediction accuracy under the format of predictions, which are pre-sigmoid values (in range $[-\inf, +\inf]$) by default, by wrapping the sigmoid transformation and accuracy checking together. In comparison, function \texttt{score\_eval\_func()} is the method to return metrics other than accuracy. In label-imbalanced binary classification, accuracy cannot reliably reveal the performance quality on its own as the metric can be 'tricked' by predicting all the instances as the majority class. This type of prediction will lead to high accuracy, yet the classifier actually does nothing. Thus, metrics taking 'preciseness' into accounts, such as precision, recall, $F_1$ score and Matthew's Correlation Coefficient (MCC) (\cite{matthews1975comparison}), are often applied for the scenario (\cite{powers2011evaluation}). Function \texttt{score\_eval\_func()} provides implementations for all the metrics mentioned above, and it can be overloaded by specifying the partial argument 'mode'(which can be accomplish by \texttt{functools.partial()}).  \par

\section{Theories and Derivatives}
\label{sec:derivatives}
In this section, the mathematical foundations  and derivations for the loss functions are discussed. For a high-level introduction, since XGBoost adopts an additive learning scheme with a second-order approximation, the first-order derivative (short-handed as 'gradient') and second-order derivative (noted as 'hessian' although somehow a misnomer) of the loss functions with respect to the prediction are required for fitting the model. To illustrate a clear mechanism, the section will first review the second-order approximation of additive tree boosting in section \ref{subsec:secondoderapprox}. Subsequently, the derivatives of gradients and hessians of the weighted and focal losses will be discussed in section \ref{subsec:weightedloss} and \ref{subsec:focalloss}, respectively. \par
The notations used in this section will be as follows. We use $m$ to denote the number of data and $n$ for the number of features. The 'raw prediction' before the sigmoid function will be denoted as $z_{i}$, and the probabilistic prediction will be $\hat{y_{i}}=\sigma(z_{i})$, where $\sigma(\cdot)$ represents the Sigmoid function. It is important to keep in mind that there is a discrepancy between the notations of this paper and the original XGBoost literature (\cite{chen2016xgboost}), as the $\hat{y}$ in their analysis is denoted as $z$ here. $y_{i}$ is used to denote the true label, and $\alpha$ and $\gamma$ are used for the parameters for the two loss functions, respectively.
\subsection{Second-order Approximation of Gradient Boosting Tree}
\label{subsec:secondoderapprox}
Denoting the input as $\boldsymbol{x}_{i}$ and the target as $y_{i}$, according to \cite{chen2016xgboost}, the additive learning objective used in practice is:\\
\begin{equation}
\label{equ:originaladditive}
\mathcal{L}^{(t)} = \sum_{i=1}^{n}l(y_{i},z_{i}^{(t-1)}+f_{t}(\boldsymbol{x}_{i})) + \Omega(f_{t})
\end{equation}
Equation \ref{equ:originaladditive} is a \emph{regularized objective}, where $l(\cdot, \cdot)$ denotes the loss between the targets ($y_{i}$) and predictions ($z_{i}$), and $\Omega(\cdot)$ denotes the complexity of the model. The tree is fitted in an additive manner with $z_{i}^{(t)}=z_{i}^{(t-1)}+f_{t}(\boldsymbol{x}_{i})$, and $t$ stands for the $t$-th iteration of the training process. Notice that the notations used above are slightly different from the original literature. Our goal is to find the $f_{t}(\cdot)$ of the $t$-th iteration to optimize objective \ref{equ:originaladditive}; To achieve this, applying a second-order Taylor expansion on equation \ref{equ:originaladditive}, one will get:\\
\begin{equation}
\label{equ:approxadditive}
\begin{split}
\mathcal{L}^{(t)} &\approx \sum_{i=1}^{n}[l(y_{i},z_{i}^{(t-1)})+g_{i}f_{t}(\boldsymbol{x}_{i})+\frac{1}{2}h_{i}(f_{t}(\boldsymbol{x}_{i}))^{2}] + \Omega(f_{t})\\
& \propto \sum_{i=1}^{n}[g_{i}f_{t}(\boldsymbol{x}_{i})+\frac{1}{2}h_{i}[f_{t}(\boldsymbol{x}_{i})]^{2}] + \Omega(f_{t})
\end{split}
\end{equation}
The last line comes from the fact that the $l(y_{i},z_{i}^{(t-1)})$ term can be removed from the learning objective as it is unrelated to the fitting of the model in the $t$-th iteration. In equation \ref{equ:approxadditive}, there are $g_{i}=\frac{\partial L}{\partial z_{i}}$ and $h_{i}=\frac{\partial^{2} L}{\partial z_{i}^{2}}$, which are the 'gradient' and 'hessian' terms mentioned before. Notice that both $g_{i}$ and $h_{i}$ are scalars, as individual boosting trees only deal with binary problems. Multi-class classification tasks are usually processed by an ensemble of binary classification trees (so-called \emph{one-vs-all} scheme, see \cite{allwein2000reducing,eibl2005multiclass}).\par
Since XGBoost does not provide automatic differentiation, the hand-derived derivatives will be essential. Meanwhile, the derived expressions have further potentials to be applied into other machine learning tasks. Therefore, the derivatives are discussed in sections \ref{subsec:weightedloss} and \ref{subsec:focalloss}. For both loss functions, sigmoid is selected as activation, and the following basic property of sigmoid will be consistently used in the derivatives:
\begin{equation}
\label{equ:sigmoidderive}
\begin{split}
\frac{\partial \hat{y}}{\partial z} &= \frac{\partial \sigma(z)}{\partial z} \\
& = \sigma(z)(1-\sigma(z))\\
& = \hat{y}(1-\hat{y})
\end{split}
\end{equation}

\subsection{Weighted Cross-entropy Loss}
\label{subsec:weightedloss}
The weighted cross-entropy loss for binary classification can be denoted as follows: 
\begin{equation}
L_{w}=-\sum_{i=1}^{m}\big(\alpha y_{i}\log(\hat{y}_{i})+(1-y_{i})\log(1-\hat{y}_{i})\big)
\end{equation}
where $\alpha$ indicates the 'imbalance parameter'. Intuitively, if $\alpha$ is greater than 1, extra loss will be counted on 'classifying 1 as 0'; On the other hand, if $\alpha$ is less than 1, the loss function will weight relatively more on whether data points with label 0 are correctly identified. \par
The first order derivative is presented as follows:\\
\begin{equation}
    \frac{\partial L_{w}}{\partial z_{i}} = -\alpha^{y_{i}}(y_{i}-\hat{y}_{i})
\end{equation}
The derivative is similar with the $\partial L/\partial z$ term for ordinary cross-entropy loss. A notable difference is that a $\alpha^{y_{i}}$ term is added to control the present of the parameter. \par
Taking derivative with respect to $z_{i}$ again, one will get the second-order derivative:\\
\begin{equation}
    \frac{\partial L_{w}^{2}}{\partial^{2} z_{i}} = \alpha^{y_{i}}(1-\hat{y}_{i})(\hat{y}_{i})
\end{equation}
After plugging equation \ref{equ:sigmoidderive} to the derivation.
\subsection{Focal Loss}
\label{subsec:focalloss}
According to \cite{lin2017focal}, the binary focal loss can be denoted as:
\begin{equation}
    L_{f} = -\sum_{i=1}^{m}y_{i}(1-\hat{y}_{i})^{\gamma}\log(\hat{y}_{i}) + (1-y_{i})\hat{y}_{i}^{\gamma}\log(1-\hat{y}_{i})
\end{equation}
As one can observe, if one sets $\gamma=0$, the equation will become ordinary cross-entropy loss. Taking equation \ref{equ:sigmoidderive} into consideration, the first derivative of the focal loss can be denoted as:
\begin{equation}
\label{equ:firstorderfocal}
\begin{split}
    \frac{\partial L_{f}}{\partial z_i} = & \gamma[(\hat{y}_i+y_{i}-1)(y_{i}+(-1)^{y_{i}}\hat{y}_{i})^{\gamma}\log(1-y_{i}-(-1)^{y_{i}}\hat{y}_{i})]\\
    & + (-1)^{y_i}(y_{i}+(-1)^{y_{i}}\hat{y}_{i})^{\gamma+1}
\end{split}
\end{equation}
And if $\gamma$ is set to 0 in equation \ref{equ:firstorderfocal}, the derivative will be the same as cross-entropy loss. The equation follows a clear structure, but it is still lengthy. To simplify the expression, one can set the following short-hand variables:
\begin{equation}
    \begin{cases} 
    \eta_1 = \hat{y}_{i}(1-\hat{y}_{i}) \\ 
    \eta_2= y_{i} + (-1)^{y_{i}}\hat{y}_{i}\\ 
    \eta_3 = \hat{y}_i + y_{i}-1\\ 
    \eta_4 = 1-y_{i}-(-1)^{y_{i}}\hat{y}_i
    \end{cases}
\end{equation}
Plugging these representations into equation \ref{equ:firstorderfocal}, the first-order derivative can be denoted as:
\begin{equation}
\label{equ:shorthandfocal}
    \frac{\partial L_{f}}{\partial z_i} = \gamma \eta_3 \eta_2^{\gamma} \text{log}(\eta_4) + (-1)^{y_{i}}\eta_2^{\gamma + 1}
\end{equation}
Finally, taking derivatives with respect to $z_{i}$, and combining with equation \ref{equ:sigmoidderive} and \ref{equ:shorthandfocal}, one can get the second-order derivative ('hessian'), which can be denoted as:
\begin{equation}
\label{equ:secondorderfocal}
\begin{split}
    \frac{\partial^{2} L_{f}}{\partial z_{i}^{2}} =  \eta_{1}&\{\gamma[(\eta_2^{\gamma}+\gamma (-1)^{y_{i}}\eta_3 \eta_2^{\gamma - 1})\text{log}(\eta_4)-\frac{(-1)^{y_i}\eta_3 \eta_2^{\gamma}}{\eta_4}]\\
    & + (\gamma+1)\eta_2^{\gamma}\}
\end{split}
\end{equation}
Again, if $\gamma=0$, the second-order derivative becomes $\eta_{1}=\hat{y}_{i}(1-\hat{y}_{i})$, which matches the formula of the second-order derivative of ordinary cross-entropy.

\section{Related Work}
\label{sec:literaturereview}
The paper is built on the foundation of the original papers of XGBoost (\cite{chen2016xgboost}) and focal loss (\cite{lin2017focal}), and the methodology to program customized loss function is provided in the software's Github page\footnote[1]{\url{https://github.com/dmlc/xgboost/blob/master/demo/guide-python/custom_objective.py}}. XGBoost is based on the algorithm of gradient tree boosting (\cite{friedman2001greedy}), and this method has been deemed as a powerful Machine Learning technique long before the XGBoost was born (\cite{natekin2013gradient}). Besides XGBoost, there are other implementations of gradient boosting, such as pGBRT (\cite{tyree2011parallel}), LightGBM (\cite{ke2017lightgbm}), and CatBoost (\cite{prokhorenkova2018catboost}). Some of the implementations have additional features and are able to outperform XGBoost on some specific problems, but XGBoost remains the method-of-the-first-choice in the data science community at large. As for the recently proposed focal loss, studies related to it are usually affiliated with Neural Networks and Deep Learning (\cite{wang2018vehicle,demir2018deepglobe,zhang2019objectdetect}). The loss function is usually applied in an end-to-end manner with automatic differentiation, and to the best of our knowledge, there has not been any notable publication comprehensively discussing the derivatives of the loss function (despite the first-order derivative was briefly discussed in another form in the original paper).\par
Previous applications of XGBoost in label-imbalanced scenarios focus mostly on data-level algorithms. For example, \cite{kabir2018resampling} applies several commonly-used data resampling methods before using XGBoost for label-imbalanced breast cancer classification, and \cite{he2018novel} utilized a more advanced under-sampling method called BalanceCascade (\cite{liu2008exploratory}) with XGBoost for credit scoring. Among the limited number of publications discussing algorithm-level modification for XGBoost in imbalanced classification, \cite{xia2017cost} used a \textit{a prior} modification of the sigmoid activation to achieve a better result, but the loss function was unchanged. As it has been mentioned in section \ref{sec:introduction}, \cite{chen2017radar} is by far the only implementation explicitly applied weighted function to XGBoost to best of our knowledge. It is noticeable that a Tensorflow-based gradient boosting implementation called \textit{Tf Boosted Trees} (\cite{ponomareva2017tfboosting}) is able to run with the loss functions without the derivatives provided in this paper as it has an automatic differentiation mechanism. Nevertheless, it is a less popular package without supports of large-scale Machine Learning and compatibility with \textit{Scikit-learn} toolkit. \par
As a common issue frequently encountered in practice, label-imbalanced classification has been intensively studied by researchers and there are multiple existing software programs designed to handle the problem. For a great example, \cite{lemaitre2017imbalanced} provides an integrated Python package called \textit{Imbalanced-learn} for data-level resampling for imbalanced classification, and it has similar counterparts in the regime of other programming languages, such as \textit{ROSE} in R (\cite{lunardon2014rose}). It is worth noting that the \textit{Imbalanced-learn} package can be considered as an extension of \textit{Scikit-learn}, and the Machine Learning toolkit itself also provides elementary methods to deal with label-imbalanced problems (\cite{pedregosa2011sklearn}). Other software programs concerning label-imbalanced classification include popular Data Mining toolkits, such as KEEL (\cite{alcala2011keel}) and WEKA (\cite{hall2009weka}). In addition, \cite{zhang2019multi} provides a software containing a set of algorithms specifically for multi-class label-imbalanced problems, serving as one of the most recent studies on this topic.

\section{Experiments}
\label{sec:experiment}
In this section, experimental results of the proposed Imbalance-XGBoost package are demonstrated in two parts: (1) Binary classification on Parkinson Disease data, a recently-collected imbalance dataset; (2) Binary classification on four benchmark datasets (with imbalance ratio) published in UCI Machine Learning Repository (\cite{Dua:2019}): ecoli(9:1), arrhythmia(17:1), car\_eval\_4(26:1) and ozone(42:1). To provide an observation of the performance improvement particularly by two loss functions, we compare the results of Weighted-XGBoost and Focal-XGBoost with vanilla XGBoost using the same parameter set, and the performances are also compared against SVM and Logistic regression.

\subsection{Parkinson Disease Dataset and Setup}
\label{subsec:dataset}

The Parkinson's Disease(PD) classification data\footnote{available publicly, url: \url{https://archive.ics.uci.edu/ml/datasets/Parkinson\%27s+Disease+Classification}} is a recently introduced dataset (\cite{sakar2019parkinson}) with 757 features categorized into 7 specific groups (two originally separate groups, Bandwidth and Formant, are merged in experiments), the data was gathered from 188 Parkinson's disease patients and 64 healthy individuals at the Department of Neurology in Cerrahpaşa Faculty of Medicine, Istanbul University. Each individual corresponds to 3 records, and due to the differences between the number of participants of the two sides, a label imbalance ratio of 188:64 (roughly 3:1) emerges. \par

The original literature \cite{sakar2019parkinson}  demonstrated the classification performances with seven conventional classification algorithms and two ensemble approaches. The experiments in the original literature were conducted with leave-one-object-out cross validation. To keep consistent with the original system, the same cross validation setup is applied in this paper. Furthermore, the results in \cite{sakar2019parkinson} illustrate a high accuracy and relatively lower $F_1$ score, indicating that the classifiers failed to tell the two classes clearly and likely achieved the performance by overwhelming predicting the majority class. This is an unfavourable behavior in label-imbalanced classification, and one advantage of \textit{Imbalance-XGBoost} is that it does not suffer from this problem. This characteristic can be further verified by the ROC and Precesion-Recall (PR) curves in figures 2-4.\par
To conduct the leave-one-object-out cross validation, the correctness collection function mentioned in section \ref{subsec:builtinscore} is applied. By collecting results of True\_Positive (TP), True\_Negative (TN), False\_Positive (FP), and False\_Negative (FN), the confusion metric can be obtained and accuracy and $F_1$ score are computed accordingly. The records are evaluated in a \textit{per-record} manner, which means the 3 records of one object (patient/healthy individual) will be evaluated individually, and 3 counts of the correctness will be added. 

\subsection{Classification Results and Discussion}
\label{subsec:experimentresults}
Accuracy and $F_1$ score of the test set with 6 sets of features are presented in Table \ref{tab:individualfeature}, where Best in \cite{sakar2019parkinson} indicates the best performance of accuracy and $F_1$ score retrieved from the paper.\par
\begin{table}[!h]
\centering
\caption{Classification performance on individual features}\label{tab:individualfeature}
\resizebox{0.6\textwidth}{!}{
\begin{tabular}{lllll}
\toprule
 & \multicolumn{2}{l}{Baseline features} & \multicolumn{2}{l}{MFCC}\\
\midrule
 & Accuracy & $F_1$ score & Accuracy & $F_1$ score\\
Best in \cite{sakar2019parkinson} & 0.79 & 0.75 & 0.84 & 0.83\\
XGBoost & 0.76 & 0.84 & 0.80 & 0.87\\
Weighted-XGBoost & 0.76 & \textbf{0.85} & 0.80 & 0.87\\
Focal-XGBoost & 0.76 & \textbf{0.85} & 0.82 & \textbf{0.89}\\
\midrule
 & \multicolumn{2}{l}{Wavelet Features} & \multicolumn{2}{l}{Bandwidth + Formant}  \\
\midrule
 & Accuracy & $F_1$ score & Accuracy & $F_1$ score\\
Best in \cite{sakar2019parkinson} & 0.78 & 0.74 & 0.77 & 0.72  \\
XGBoost & 0.72 & 0.82 & 0.71 & 0.83\\
Weighted-XGBoost & 0.75 & \textbf{0.85} & 0.74 & \textbf{0.85}\\
Focal-XGBoost & 0.75 & \textbf{0.85} & 0.75 & \textbf{0.85}\\
\midrule
 & \multicolumn{2}{l}{Intensity Based} & \multicolumn{2}{l}{Vocal Fold-Based}\\
\midrule
 & Accuracy & $F_1$ score & Accuracy & $F_1$ score\\
Best in \cite{sakar2019parkinson} & 0.77 & 0.74 & 0.77 & 0.74\\
XGBoost & 0.73 & 0.83 & 0.72 & 0.82\\
Weighted-XGBoost & 0.75 & \textbf{0.85} & 0.75 & 0.84\\
Focal-XGBoost & 0.75 & \textbf{0.85} & 0.76 & \textbf{0.85} \\
\bottomrule
\end{tabular}
}
\end{table}
\par

Without exception, although a slight declination of accuracy could be observed in weighted-XGBoost and focal-XGBoost, both classifiers generate a significantly higher $F_1$ score. The increase of $F_1$ score and the decrements of accuracy suggest that the previous-obtained higher accuracy is a consequence of overlooking minority class. Furthermore, for almost all the feature groups, the highest $F_1$ score is obtained by focal-XGBoost. This observation can be explained from an algorithmic perspective that focal loss is more robust to parameters, while weighted loss is prone to the effect of sub-optimal parameters even if parameter search is applied.\par
\cite{sakar2019parkinson} also applied a classifier with 50 top-ranked features selected by mRMR (minimum Redundancy-Maximum Relevance). The feature selection technique is based on the principle of maximizing the joint dependency of top ranking variables on the targeted one by reducing the redundancy among them (\cite{peng2005feature,garcia2016tutorial}). For a comparison purpose, this paper employs the same technique with provided Python interface\footnote[1]{\url{https://github.com/fbrundu/pymrmr}}, and produces a subset of top-50 features to run with \textit{Imbalance-XGBoost}. The performance of weighted- and focal-XGBoost on the top-50 features can be observed in table \ref{tab:top50feature}.

\begin{table}[!h]
\centering
\caption{Classification performance on top 50 features}\label{tab:top50feature}
\begin{tabular}{llll}
\toprule
 & Top 50 Features \\
\midrule
 & Accuracy & $F_1$ score \\
Best in \cite{sakar2019parkinson} & 0.86 & 0.84 \\
XGBoost & 0.83 & 0.89  \\
Weighted-XGBoost & 0.82 & 0.88 \\
Focal-XGBoost & 0.84 & \textbf{0.90} \\
\bottomrule
\end{tabular}
\end{table}
\par
Consistent with the performance on individual groups of features, focal-XGBoost classifier has the highest $F_1$ score, slightly better than weighted-XGBoost. Both weighted- and focal-XGBoost outperform best classifier in \cite{sakar2019parkinson} by a large margin, and since the top-50 feature can be regarded as a 'master subset', the superiority of the methods implemented in imbalance-XGBoost can be further corroborated. \par

\begin{figure*}[!h]
\centering
\includegraphics[width=0.8\textwidth]{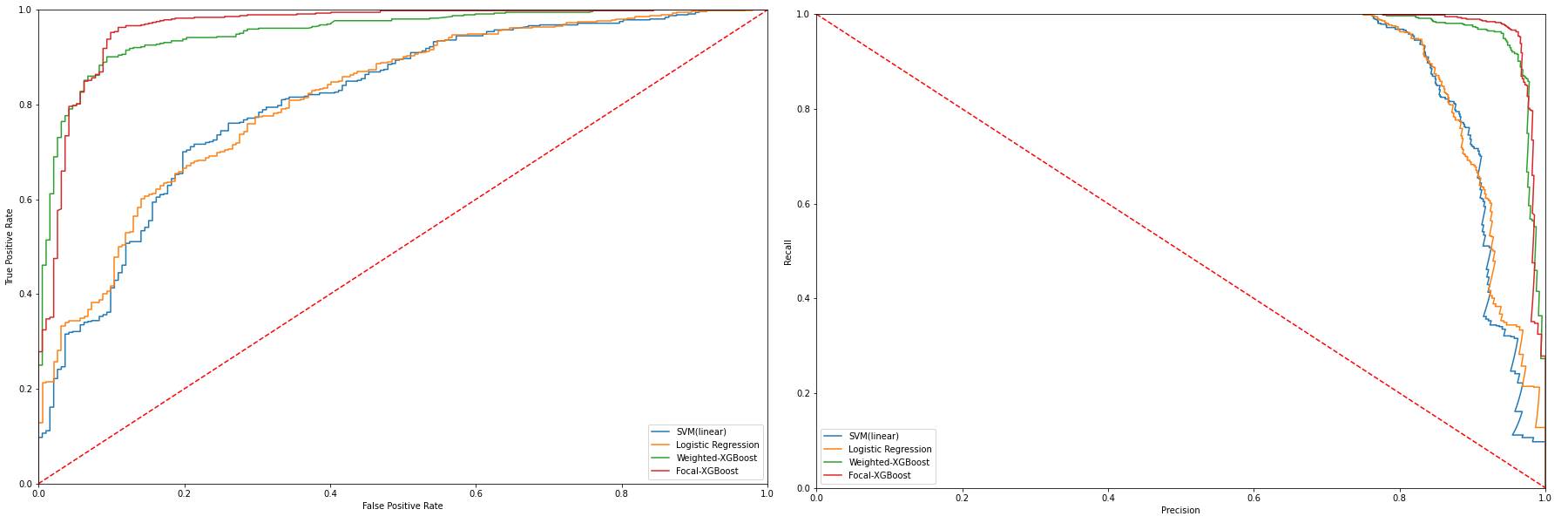}
\caption{\label{fig:baseline}ROC and Precision-Recall Curve of Baseline Feature}
\end{figure*}

\begin{figure*}[!h]
\centering
\includegraphics[width=0.8\textwidth]{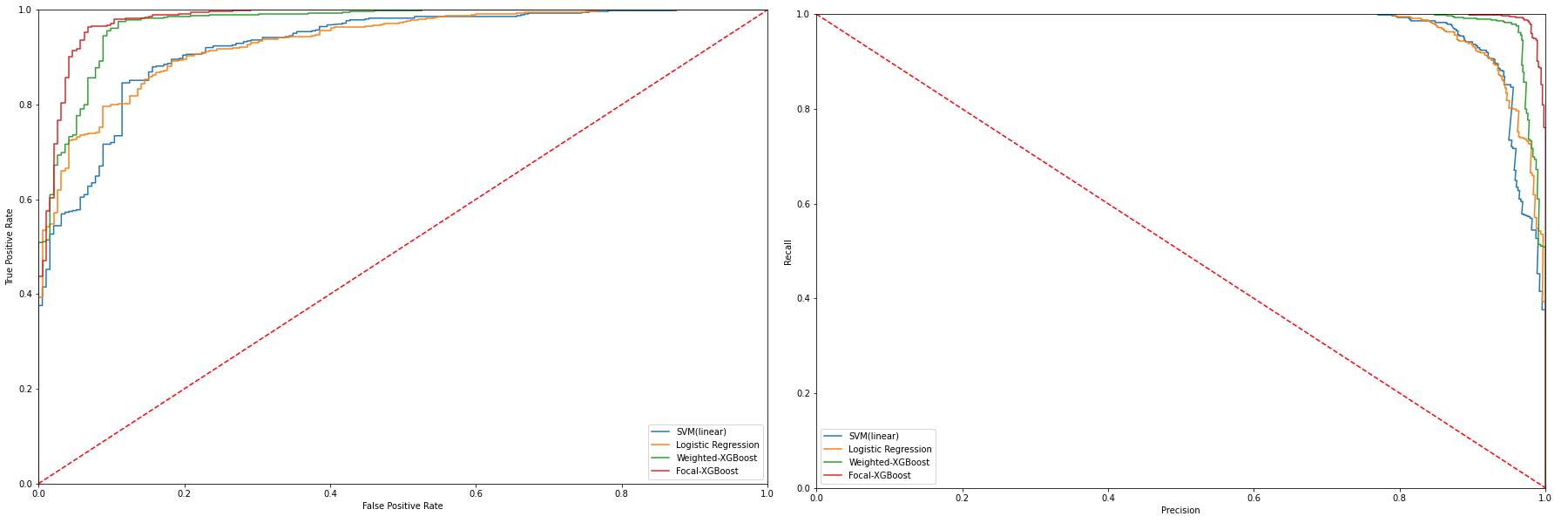}
\caption{ROC and Precision-Recall Curve of MFCC Feature}
\label{fig:mfcc}
\end{figure*}

\begin{figure*}[!h]
\centering
\includegraphics[width=0.8\textwidth]{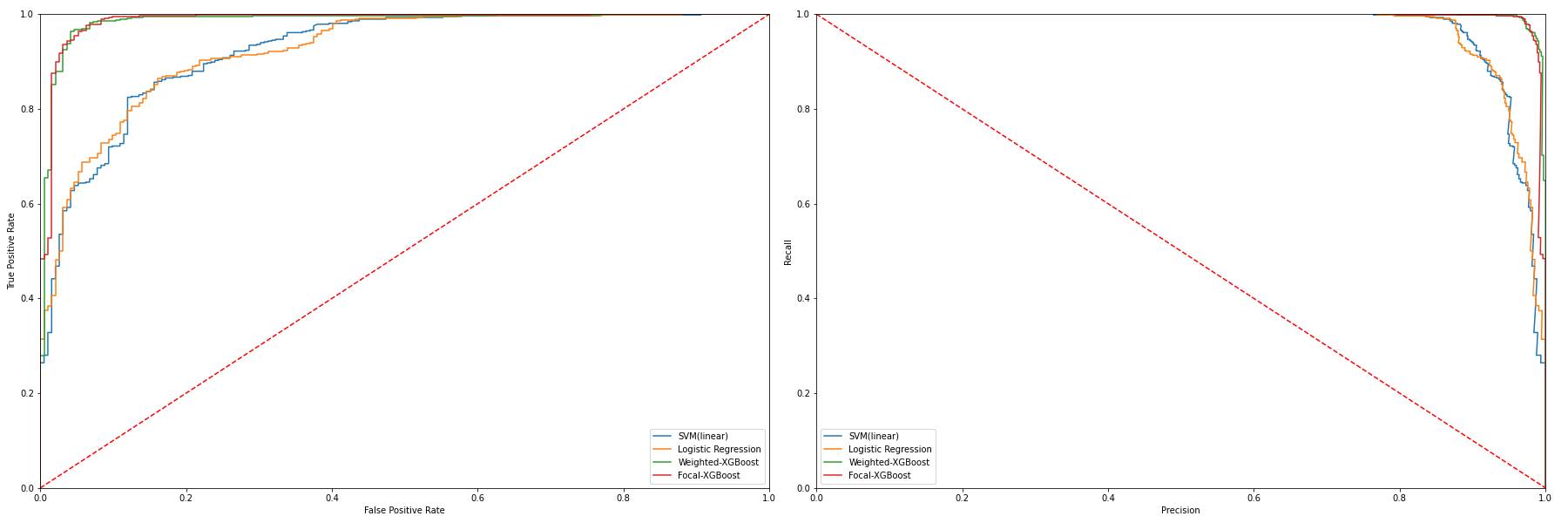}
\caption{ROC and Precison-Recall Curve of Top50 Features}
\label{fig:top50}
\end{figure*}

To provide further insights for the predictions of weight- and focal-XGBoosts, the ROC and Precision-Recall (PR) curves of the two boosting trees and two other methods (SVM and Logistic Regression, which are used in \cite{sakar2019parkinson}) are illustrated in figures \ref{fig:baseline}-\ref{fig:top50}. Limited by the space, only the feature sets of Baseline, MFCC and Top 50 are adopted. From the figures, it can be observed that the XGBoost-based models consistently provide superior performance in terms of AUC and PR curve. This further validate our earlier assertion that the higher accuracy of the method in \cite{sakar2019parkinson} comes from mis-classifying the minority instances into the majority class.

\subsection{Classification for Benchmark Imablanced Datasets }
To further evaluate the performance of Imbalance-XGBoost, we introduce four other imbalanced datasets for binary classification from UCI Machine Learning Repository: ecoli, arrhythmia, car\_eval\_4 and ozone. Notice that the Parkinson Disease dataset has an imbalance ratio of 3:1, while these four datasets have higher ratios progressively as 9:1 (ecoli), 17:1 (arrhythmia), 26:1 (car\_eval\_4), 42:1 (ozone). Slightly different from the experiments on Parkinson Dataset, we apply K-Fold(K=5) cross validation for Grid Search instead of leave-one-out, on the grounds that these four datasets include a considerable number of instances and K-Fold is commonly used in practice. \par
Our focus here is to illustrate the effectiveness of the weighted and focal loss functions, so we compare weighted- and focal-XGBoosts againt its vanilla counterpart. The classification results are presented in table \ref{tab:benchdataset}, with accuracy, $F_1$ score and MCC(Matthews Correlation Coefficient) as evaluation metrics.

\begin{table}[!h]
\centering
\caption{Classification performance on benchmark imbalanced datasets}\label{tab:benchdataset}
\resizebox{0.7\textwidth}{!}{
\begin{tabular}{lllllll}
\toprule
 & \multicolumn{3}{l}{ecoli (9 : 1)} &   \multicolumn{3}{l}{arrhythmia (17 : 1)} \\
\midrule
 & Accuracy & $F_1$ score & MCC & Accuracy & $F_1$ score & MCC\\
XGBoost & 0.926 & 0.627 & 0.586 & 0.958 & 0.627 & 0.605\\
Weighted-XGBoost & 0.935 & 0.645 & 0.616 & 0.967 & 0.681 & 0.665\\
Focal-XGBoost & 0.935 & 0.645 & 0.616 & 0.960 & 0.679 & 0.662\\
\midrule
 & \multicolumn{3}{l}{car\_eval\_4 (26 : 1)} & \multicolumn{3}{l}{ozone (42 : 1)}  \\
\midrule
 & Accuracy & $F_1$ score & MCC & Accuracy & $F_1$ score & MCC\\
XGBoost & 0.904 & 0.239 & 0.217 & 0.967 & 0.144 & 0.154\\
Weighted-XGBoost & 0.959 & 0.268 & 0.266 & 0.959 & 0.268 & 0.266\\
Focal-XGBoost & 0.911 & 0.287 & 0.273 & 0.964 & 0.180 & 0.173 \\
\bottomrule
\end{tabular}
}
\end{table}
\par

From the table, it can be found that the classification accuracy of weighted- and focal-XGBoost is at least as competitive as XGBoost. In terms of $F_1$ score and MCC, which are considered the most crucial metrics for imbalanced classification, weighted- and focal-XGBoost outperforms vanilla XGBoost by a large margin on all datasets. We also remark that as the imbalance ratio goes up, the improvements on $F_1$ score and MCC for weighted- and focal-XGBoosts become more significant.

\section{Conclusion}
\label{sec:conclusion}
This paper presents a novel Python-based package, namely \textit{Imbalance-XGBoost}, for binary label-imbalanced classification with XGBoost. The package implemented weighted cross-entropy and focal loss functions on XGBoost, and it is fully compatible with the popular \textit{Scikit-learn} package in Python. The design and usage of the package are introduced, and the discussion of methods provide a clear and comprehensive user guidance. The theories and derivatives essential to the package are further discussed, and experiments based on five imbalance classification datasets are conducted with competitive performances illustrated. Overall, the package demonstrated in this paper successfully combines XGBoost with popular label-imbalance-robust loss functions and provides one of the most competitive performances up to date. \par
In summary, this paper has made three main contributions. Firstly, the paper has introduced a novel package that leverages the power of weighted and focal loss function for XGBoost, and it has huge potentials to be applied to a variety of real-life binary classification problems. Secondly, the paper has studied the theoretical foundations of the second-order approximation of XGBoost and has provided essential derivatives for the loss functions to be applied. The derivatives can also be applied to other fields in Machine Learning, and the equations in the merged form are convenient to be vectorized. And finally, the paper has offered new advanced performances on the five benchmark datasets, and the emphasis of the imbalanced nature provides new a perspective to study them. \par

\bibliographystyle{unsrt}  
\bibliography{references}  
\newpage

\section*{Supplementary Materials}
Two code listings regarding the usage of the \textit{Imbalance-XGBoost} package are illustrated here. Details of the explanations of the codes can be found in section \ref{sec:softwaredesign}.
\\* 
\\* 
\par
\par
\begin{minipage}{0.9\textwidth}
\begin{lstlisting}[style=mypython,basicstyle=\footnotesize,caption={Basic Usage of Imbalance-XGBoost},captionpos=b,label=lst:sampleusage]
from imxgboost.imbalance_xgb import imbalance_xgboost as imb_xgb
# weighted XGBoost
xgboster_weighted = imb_xgb(special_objective='weighted')
xgboster_weighted.fit(X, y, imbalance_alpha=2.0)
# focal XGBoost
xgboster_focal = imb_xgb(special_objective='focal')
xgboster_focal.fit(X, y, focal_gamma=2.0)
\end{lstlisting}
\end{minipage}
\\* 
\\* 
\\* 
\par
\par
\begin{minipage}{0.9\textwidth}
\begin{lstlisting}[style=mypython,basicstyle=\footnotesize,caption={Parameter Tuning and Model Evaluation of Imbalance-XGBoost with SK-learn},captionpos=b,label=lst:paratuning]
from imxgboost.imbalance_xgb import imbalance_xgboost as imb_xgb
from sklearn.model_selection import GridSearchCV, LeaveOneOut, cross_validate
# focal XGBoost
xgboster_focal = imb_xgb(special_objective='focal')
# cross-validatoin booster
CV_focal_booster = GridSearchCV(xgboster_focal, {"focal_gamma":[1.0,1.5,2.0,2.5,3.0]})
# fit the booster
CV_focal_booster.fit(X, y)
# retrieve the best model and parameter
opt_focal_booster = CV_focal_booster.best_estimator_
opt_focal_parameter = CV_focal_booster.best_params_
# instantialize an imbalance-xgboost instance for 
# cross-validation
xgboost_focal_opt = imb_xgb(special_objective='focal', **xgboost_opt_param)
# initialize a leave-one splitter
loo_splitter = LeaveOneOut()
# Leave-One cross validation
loo_info_dict = cross_validate(xgboost_focal_opt, X=x, y=y, cv=loo_splitter)
\end{lstlisting}
\end{minipage}

\end{document}